# Task Allocation in Mobile Robot Fleets: A review

*A. Meseguer Valenzuela, Universitat Politècnica de València, Spain (anmeva1@doctor.upv.es), Instituto Tecnológico de Informática (ITI), Spain (ameseguer@iti.es), ORCID: https://orcid.org/0000-0002-2328-6422*

*F. Blanes Noguera, Instituto de Automática e Informática Industrial, Universitat Politècnica de València, Spain (pblanes@ai2.upv.es), ORCID: https://orcid.org/0000-0002-9234-5377*

*Abstract.* Mobile robot fleets are currently used in different scenarios such as medical environments or logistics. The management of these systems provides different challenges that vary from the control of the movement of each robot to the allocation of tasks to be performed. Task Allocation (TA) problem is a key topic for the proper management of mobile robot fleets to ensure the minimization of energy consumption and quantity of necessary robots. Solutions on this aspect are essential to reach economic and environmental sustainability of robot fleets, mainly in industry applications such as warehouse logistics. The minimization of energy consumption introduces TA problem as an optimization issue which has been treated in recent studies. This work focuses on the analysis of current trends in solving TA of mobile robot fleets. Main TA optimization algorithms are presented, including novel methods based on Artificial Intelligence (AI). Additionally, this work showcases most important results extracted from simulations, including frameworks utilized for the development of the simulations. Finally, some conclusions are obtained from the analysis to target on gaps that must be treated in the future.

## I. INTRODUCTION

THE integration of Autonomous Guided Vehicles (AGV) and Autonomous Mobile Robots (AMRs) has generated a new set of applications in unknow and known environments. From emergency scenarios to industrial applications, the integration of ground-based robots is a key factor to boost the automation of their processes. In all these scenarios, an optimal management is necessary to ensure the behavior of the system. For this aim, the management must cover three main tasks. Firstly, Task Allocation (TA) where different robots receive tasks to be completed considering at least the feasibility to perform the task and the cost. Secondly, Path Planning (PP) which is focused on calculating a trajectory the robot must follow from origin point to reach a target objective. The last task named Motion Planning (MP) is mainly focused on controlling the movement of each robot to follow the trajectory previously calculated.

Considering the different tasks to be performed, TA is an important stage of robot fleet management due to it affects to the trajectories that each robot will follow, thus the distance travelled, related robot energy consumption and even the minimum quantity of robots necessary for the activities. Therefore, a proper TA algorithm aims the assignment of tasks to ensure the production but reducing the costs to ensure the economic feasibility of robot fleet applications. During recent years there has been a research trend related to TA which has been evolving with the integration of AMRs and evolution of AI. Focusing on this topic, this work introduces TA algorithms applied in ground-based robot fleet applications, providing a new update compared to state-of-the-art reviews (De Ryck et al., 2020; Meseguer, A., Blanes, F., 2023). Moreover, this study offers insights about the application of AI into the management of robot fleets to analyze the impact of this trend.

This work analyzes different research publications that are indexed in Google Scholar. Some filters were applied to extract publications that are more focused on TA in mobile robot fleet scenarios. Although the target use case is indoor logistics for industry, other applications are under the scope to observe differences within the use cases and the type of fleet management. Additional filters were applied to get the most recent publications, prioritizing works from the last two years. Therefore, from the 1440 results related to TA and robot mobile fleets that can be found in Google Scholar, this work considers 52 publications as the most relevant, which are described in the next section.

The structure of this work can be described as follows: Section II that provides an overview of the main algorithms. Section III displays different simulations and experimental scenarios that have been detected. Finally, Section IV offers last findings and conclusions.

## II. Algorithms

The development of an entity capable to make decisions to solve problems can be defined as an intelligent agent. The combination of different agents can provide solutions where just one entity cannot, achieving what we know as multiagent system (MAS). The concept, which was introduced in (Genesereth & Ketchpel, 1994) has been analyzed mainly in terms of orchestration algorithms that enhance the performance of MAS. At application level (such as robotics), the target in this management is that intelligent agents may perform different tasks at the lowest cost (energy consumption, distance travelled, number of robots...). Therefore, the orchestration issue has become into an optimization problem to minimize these costs.

There is a wide variety of applications for MAS. It can include management of power smart grids (Boussaada et al., 2016), wireless communications (Zhao et al., 2023) or robot applications (García et al., 2012), which is the main focus of this work. In this field, there are different use cases from unknown environments (e.g. emergency or search and rescue applications…) to known environments (e.g. indoor logistics, manufacturing…).

Additionally, there are multiple types of robots that can be used as agents. From Unmanned Aerial Vehicles (UAVs)(Z. Yan et al., 2023) to Autonomous Underwater Vehicles (AUVs)(Fang et al., 2022). However, this work focuses mainly on industry-based application, where ground mobile robot fleets are most utilized(Taranta et al., 2021).

In terms of ground-based vehicles, the most optimal devices to be applied into multi-agent intelligent systems are Autonomous Mobile Robots (AMRs), which are devices that use their own sensors to exercise self-navigation. This aspect provides an advantage compared to Auto-Guided Vehicles (AGVs), which require support from the infrastructure such as marks or rails to navigate. Therefore, these devices are more susceptible to provoke deadlocks or ineffective to avoid collisions. Due to these clear advantages, AGVs which have been traditionally used in industry for indoor logistics are being replaced with AMRs(Neher et al., 2022).

Regarding the management of AGVs and AMRs, where this study is focused, there are different approaches to perform TA while costs are minimized. They can be categorized depending on the location of the entity that carries out the orchestration (commonly referred as central agent or supervisor agent), providing two main architectures.

The first one is **Centralized Management,** where a central agent in continuous communication with all the agents can calculate the costs to perform the tasks, assigning them to each agent. This approach is more applicable to known environments where there are stable wireless networks deployed. This is caused by communication requirements to monitor the activity of each agent from a centralized location. The access to data from all agents provides access to optimal solutions that achieve the lowest cost to accomplish tasks. However, this architecture brings an increase in terms of computational cost with larger fleets (scalability issues).

The other architecture is named **Decentralized Management**. This approach is based on each robot acting autonomously and therefore calculating the costs to perform their own tasks. As each agent communicates decisions with other robots, this scope requires direct communication with neighbor robots. The transmissions between the different robots also provides an opportunity to share information about the environment. Therefore, this architecture provides a suitable option in unknown environments (O'Brien et al., 2023).

Nevertheless, TA applied in decentralized management achieves mainly local optimal solutions due to the lack of knowledge about data from distant robots. Despite of this drawback, this approach is more efficient in terms of computational cost in large fleets. This is a key aspect to avoid issues related to scalability. Both management approaches are represented in Fig. 1.

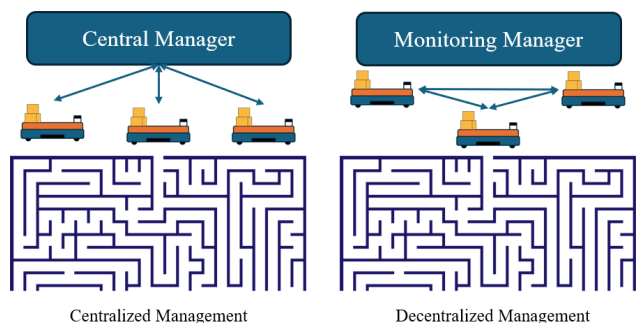

**Fig. 1.** Centralized and Decentralized Management architecture.

Regarding the different algorithms that are used to optimize TA in robot fleet applications, there are different approaches in function of the target that is desired to achieve such as exact solutions, improved scalability, constraint integration, etc... This work focuses on optimization-based algorithms which are based on mathematical expressions and finding locally or globally optimal solutions. Additionally, other types of methods are analyzed such as Market-based algorithms focused on auctioning behavior and Artificial Intelligence methods, which is a current trend in solving optimization problems such as TA. The following figure provides an overview of the methods included in this work:

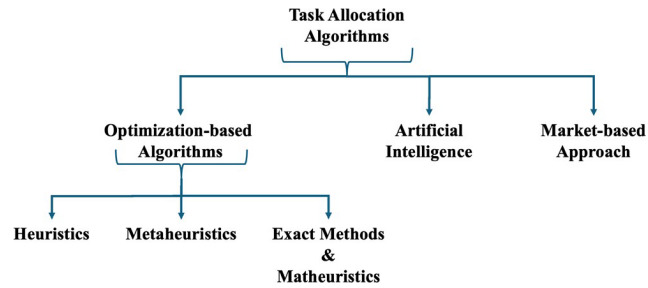

**Fig. 2.** Types of algorithms analyzed

## A. Heuristics

Considering TA methods based on using heuristic models, there are different examples that target the optimization of costs. This approach is based on achieving solutions that solve TA problem using iterative procedures mainly by applying searching methods with node graphs or trial and error applications.

In (A. Bolu & Ö. Korçak, 2021), a centralized management is proposed using heuristic model. This work assigns tasks to robots considering location of the robots and the pods to transport. There are different tasks such as charging batteries and replenishment processes. Additionally, to the optimization of the tasks, the number of pods is also controlled by a novel Order Batch to Robot Task Conversion (OBRTC).

Another example of heuristic method application is (Shi et al., 2024), where a two-stage Fulfillment-focused Simultaneous Assignment (FFSA) method is conceived in a logistics scenario. The first stage is based on a selection of critical racks that can be used by robots to accomplish all the tasks. For this aim, a Hybrid Adaptive Large Neighborhood Search (ALNS) algorithm is applied. The second stage performs simultaneous assignment of tasks and racks using Marginal-return-based assignment with candidate strategy (MRACS).

Regarding the usage of search based methods, (L. Li et al., 2023) proposes Intensive Inter-task Relationship Tree Search (IIRTS) for fast TA of heterogeneous robot fleets. This algorithm is based on Task Batch Planning Decision Tree (TB-PDT). This work considers both, logic and inter-logic constraints in a medical scenario with different type of robots envisaged for determined medical tasks (such as sterilization or nursing).

(G. S. Oliveira et al., 2022) provides another example of heuristic method application. The authors define a space decomposition-driven heuristic algorithm called Domain zone-based Capacity and Priority constrained Task Allocator (DoNe-CPTA). This method that is based on Voronoi Tesselation, targets the allocation of tasks in an environment with different constraints and heterogeneous robots.

However, in some works the algorithm selected may be based on a combination of other heuristic methods, commonly known as hyper-heuristic algorithms. (F. Yan & Di, 2023) provides an example where the management of a mobile robot fleet is targeted to reduce time cost necessary to accomplish all the tasks. The hyper-heuristic method used is based on two levels. A low-level heuristic model defined for functional task scoring, and a high-level focused on the optimization of the allocation using Particle Swarm Optimization (PSO) algorithm (traditionally considered a meta-heuristic method).

## B. Meta-heuristics

There are optimization methods that instead of finding globally optimal solutions, target the achievement of locally optimal solutions to avoid time limitations. Some of them are focused on imitating population behavior or searching applications. Some of the main efforts in meta-heuristic methods are based on Genetic Algorithms (GA). (Martin et al., 2021) provides an example of their application in a solar panel scenario with a heterogeneous robot fleet. GA are used for TA considering energy constraints of the robots. Additionally, the authors compare the solution with Branch and Bound (B&B) algorithms. The results display an improvement in efficiency of computational resources in small scenarios.

Another example of GA applied in robot fleet management is provided in (Wang & Pan, 2023). This work is focused on the development of a management solution for mobile robot fleets that combines TA and PP. The method proposed is split into three stages. Firstly, mobile robots move to the targeted pods. Secondly, the robots carry the pods to the picking station. Lastly, the robots carry the pods back to its original location. The TA problem is solved using a GA that considers the optimization of a cost function based on the travel cost calculated in "timesteps" and "flowtime".

GA can also be applied to robot fleet management in use cases different from logistics, such as multi-robot inspection missions. This use case is displayed in (Chakraa et al., 2023) using a GA to ensure the optimization of different tasks to be performed by robots with different sensors. Therefore, different tasks are assigned considering the order of the tasks' execution and sensor sharing information.

(Berndt et al., 2021) provides an alternative to GA, using different approaches such as First In-First Out (FiFo), Earliest Deadline

First (EDL) and Least Slack Time (LST) methods. This work that compares the different scopes, is targeted to allow re-allocation of tasks during the execution considering task priority values.

Alternatively, there are meta-heuristics algorithms that are based on biological behavior, also known as bioinspired algorithms. For example, (J. Qiu et al., 2023) provides a Graph Convolutional Network (GCN) method based on Ant Colony Optimization (ACO) denoted as GCN-ACO which is formulated using the Travelling Salesman Problems (TSP) and Vehicle Routing Problem (VRP). Therefore, the method proposed is based on two stages. Firstly, a heatmap that is produced to show the probability of each edge to belong to the most optimal route into the graph. Secondly, the heatmap is integrated into ACO to guide the ant colony to select edges, thus selecting the tasks considering the cost of the routes.

## C. Exact and matheuristic Methods

There are different mathematical expressions that are used to solve optimization in large systems with different constraints. These expressions, which target the achievement of global and optimal solutions are known as Exact Methods. Between the different exact optimization algorithms that have been developed, Mixed Integer Linear Programming (MILP) provides a method used in different research works with robot fleets and different constraints.

In (Akkaya & Gökçe, 2022), a MILP based model is proposed to solve a problem of TA and routing, considering battery constraints into a job-shop AGV fleet management use case. Additionally, this work that also considers partial recharge of the robots, provides a simulation test that displays a reduction of energy consumption.

(Boccia et al., 2023) provides another example of MILP based model for solving AGV scheduling problem with battery constraints (ASP-BC) defined in (Masone et al., 2021). In this case, this solution treats the bottleneck generalized problem, considering charging operations as special tasks. These exercises must be performed to ensure the robots to have sufficient battery level to complete their tasks. For this aim, the scenario analyzed includes a set of charging stations located in a warehouse. The performance of the MILP model is compared with a three step matheuristic solution (3S-MHA). The comparison shows up that MILP solution offers effective results in medium size instances, although it suffers from scalability problems compared to 3S-MHA.

Another publication that offers insights into the introduction of MILP in mobile robot fleet applications is (Hu et al., 2023). This work provides an example of MILP usage in heterogeneous AGV fleet management. The authors describe a solution based on decomposing the problem into two hierarchical levels to solve TA and PP. For this purpose, they provide a MILP based solution that gets optimization of the costs. Later, a Hybrid Discrete State Transition Algorithm (HDSTA) is applied during the execution with Tabu list method to solve dynamic scheduling.

Regarding further comparison of MILP with other Exact Methods, (Singh et al., 2022) provides the definition of a model based on MILP which is compared to a heuristic that uses ALNS and Linear Programming. Both algorithms are applied into an AGV fleet management system for TA, reducing the costs provided by tardiness requests. The comparison displays that MILP is not suitable for small-scale problems (up to 22 task requests).

Other type of method used for improved TA is Fuzzy optimization. In (Valero et al., 2023), Bellman-Zadeh fuzzy optimization technique is used to provide a new TA approach based on response-threshold methods with time constraints and models stimuli using fuzzy sets. This work is based on a decentralized architecture where each robot decides the best task to perform through the fuzzy optimization and a communication mechanism.

Regarding Exact Methods applied in TA considering energy consumption, Kuhn-Munkers provides an example of application on this field. Also known as the Hungarian Algorithm, this method is applied in (Chatzisavvas et al., 2022). As specified, this work is based on achieving a matrix of costs calculated by each robot to complete different tasks. Therefore, this algorithm is applied to get the minimum costs per task.

Similarly, the same method is used in (Msala et al., 2023) together with K-means algorithm and GA to provide a method with three stages that reduces the costs. Firstly, providing a task assignment that directs the robots to their tasks. Secondly, the set of tasks assigned is modified to optimize the pre-allocation. Finally, the allocation is re-evaluated to detect improvements. The Hungarian Algorithm is also used in (A. Samiei & L. Sun, 2024) using a novel approach defined as Distributed Matching-by-clone Hungarian-based algorithm (DM-CHBA). This method consists of two phases. Firstly, a communication step where the costs are transferred between the agents. Secondly, an assignment stage to allocate the tasks.

There are alternative matheuristic algorithms applied in robot fleet management such as Mixed Integer Quadratic Problem (MIQP) models. In (Bergmann et al., 2021), a MIQP load balancing model is defined to provide AGV-zone assignments. This activity ensures the supply of materials between station clusters by assigning different AGVs. Additionally, this logistic scenario includes a refinement loop mechanism to update the assignment of AGVs.

## D. Market-based Approach (MBA)

The are methods applied for TA in robot fleet scenarios that are based on emulation of market activities such as auctions. In (De Ryck et al., 2021), there is a TA example about how an auction can be applied. The method proposed enhances the assignment considering resource constraints and reducing the time invested in charging the robots. Another important aspect of this decentralized method is the capability to scale the size of the fleet without reducing the performance of the management. Lately, the authors enhance the performance in (De Ryck et al., 2022). This update includes ant-colony algorithms to consider routing constraints during the optimization of the TA performed by the auction activity.

Another example of MBA applications is (Tavares et al., 2023), where a MBA distributed architecture for a dynamic factory environment is presented. This method provides a reduction of time and total distance traveled by the agents using an auction procedure. During this activity, each agent offers the lowest bid as possible except if the robot is already performing a task. The authors remark in this case the importance of the communication to perform a satisfactory auction process.

(Teck et al., 2023) provides a similar approach but considering different agents such as picking stations, order agents, pod agents and vehicle agents. The allocation of tasks is performed using dispatching rules through a manager agent that employs three different methods, which are compared. Firstly, a decentral Sequential Single-Item Auction (SSIA) method. Secondly, a greedy look-ahead heuristic algorithm based on a decision tree. This method attempts to predict the impact of selecting determined tasks in terms of travelled distance. In the end, a third method defined as a regret-based task selection process. In this work, the auction method is outperformed by the other in terms of scalability and optimization of traveled distance.

All these works provide MBA methods that are executed automatically. However there are similar approaches such as in (Galati et al., 2023), that perform auction activities being supervised by a human. Following this scope, once the auction process is completed, it is necessary the approval of a human supervisor to accept the allocation of tasks.

## E. Artificial Intelligence (AI)

In the recent years, AI has been introduced in different technical fields, including robotics. Robot fleet management is one of the topics where AI is targeted to improve state-of-the-art. On this matter, there are different types of application depending on the type of AI algorithm applied. (Lei et al., 2023) provides an example of robot fleet management that fuses task decomposition, allocation, sub-task allocation and PP. The method proposed firstly divides the robot fleet into a variety of teams, using convex optimization to minimize the distance between teams, robots and their objectives. Robot location is used to place the robots at the correct regions of space by applying graph-based Delaunay triangulation method. However, the key allocation aspect is the application of self-organizing map-based neural network (SOMNN) to achieve dynamical sub-task allocation.

An alternative method to apply AI in robot fleet management is Reinforcement Learning (RL). In this scope, AI algorithms are developed to provide learning capabilities to the control or supervisor agent. Following this approach, a RL algorithm tries different values for a defined process, such as a task assignment in logistics, obtaining good and bad results, for example in terms of distance and energy optimization. The key aspect is that RL algorithms learn how to correctly manage the scenario obtaining a policy or a set of rules. (Shibata et al., 2023) provides an example of application using RL. In this work, the authors introduce a policy model defined with dynamic task priorities using global communication. This approach also utilizes a neural network-based distributed policy model to achieve timing for coordination. The distributed policy focuses on local observation to get consensus about what actions must be selected. Once the selection is complete, RL is applied using trial and error.

Another example of RL applied in TA is (Perumaal Subramanian & Kumar Chandrasekar, 2024), where Simultaneous Allocation and Sequencing of Orders Reinforcement Learning (SASORL) algorithm is proposed. In this work, three sets are defined to manage the learning of the algorithm: state, action and reward/penalty. The target is to define the distance traveled as a penalty, to ensure that the rules optimize the cost of the actions or tasks to be performed. The proposed method requires a few iterations to learn about the behavior of the process before starting to achieve optimal results. Finally, (Malus et al., 2020) provides a similar approach using AMRs by achieving a decentralized learning method that focuses on rewarding the robots to get the best optimal solutions.

## III. Simulations

Considering the different optimization methods that have been described in this work, it is necessary to display the main aspects to analyze the status of the art. On this aim, the type of the system optimized is pointed out, considering if it follows or not a centralized management. Furthermore, the overview also provides information about the framework used to develop the proposed algorithm. Another aspect that is also remarked is the size of the fleet by splitting the type robots into AGVs, AMRs or agents (which can be any type of autonomous system such as drones and robots). The last aspect remarked is the main result that can be extracted from each work.

First, most important publications that used Heuristic methods are showed in Table I. As it can be observed, all of them are focused on centralized management, mainly due to the target of searching the most optimal solution at global level. Regarding the type of algorithm applied, there is a wide variety of Heuristic methods. This aspect is provided by the inherent simplification on search-based methods and the research background. Related to the framework used for the development of the algorithms, Matlab has provided a suitable environment for two out of five publications. Software programming has also been influenced by the analytical approach of the different works. With reference to the size of the fleets, simulations are executed with up to 25 robots.

Although there is an exception related to (Shi et al., 2024), which is based on realistic datasets with up to 1923 mobile racks. In terms of simulation results, the different solutions provide reductions of completion times and route costs.

TABLE I
HEURISTIC ALGORITHMS OVERVIEW

|  | Method | Centralized / Decentralized | Framework | Fleet Size | Remarkable results |
|---|---|---|---|---|---|
| (A. Bolu & Ö. Korçak, 2021) | OBRTC | Centralized | Web-based central software | ≤ 20 AMRs, 4 picking stations and 5000 orders | Logistics. Reduction of 46% of completion times in environments with 3 pick stations |
| (Shi et al., 2024) | ALNS | Centralized | MATLAB | Realistic data based on scenario with 8 picking stations, 1923 mobile racks and 26791 products. | Displays a reduction of 45.18% of racks visits, providing an improvement on picking efficiency. |
| (L. Li et al., 2023) | IIRTS | Centralized | MATLAB and Gazebo | Simulations with 2 setups. Firstly, with up to 45 AMRs (3 different types). Secondly, 6 AMRs | IIRTS shows improvements in terms of efficiency and effectiveness |
| (G. S. Oliveira et al., 2022) | Space decomposition-driven heuristic | Centralized | C++ | 25 AMRs (3 different types) | Reduction of 45% in cost of routes and 24% in fulfillment time |
| (F. Yan & Di, 2023) | Two-level heuristic model with PSO | Centralized | Java | ≤ 10 AMRs | Better results in scenarios with large number of tasks compared to meta-heuristic and greedy algorithm |

In relation to Meta-Heuristic methods, papers analyzed provide centralized management solutions. This architecture homogeneity is caused by the idiosyncrasy of Meta-Heuristic methods, which continuously look for optimal solutions from a global perspective. However, in this case there is a common type of algorithm (GA) which represents a population-base method. In terms of framework applied for the development of the simulations, Python provides the most used language. Regarding the size of the robot fleets utilized, these works are based on smaller fleets with up to 6 robots. However, (J. Qiu et al., 2023) provides a different type of simulation based on 20 software agents. Considering the main results from the reviewed papers, different achievements are presented in terms of success rate and calculation of optimal solutions compared to other algorithms. These values are represented in Table II.

TABLE II
META-HEURISTIC ALGORITHMS OVERVIEW

|  | Method | Centralized / Decentralized | Framework | Fleet Size | Remarkable results |
|---|---|---|---|---|---|
| (Martin et al., 2021) | Genetic and B&B Algorithm | Centralized | Montecarlo Simulation (language not defined) | ≤ 4 agents | B&B algorithm gets optimal solution. GA gets suboptimal solution in a shorter period of time |
| (Wang & Pan, 2023) | Genetic Algorithm | Centralized | MATLAB | 1 picking station, 3 AGVs and 5 different tasks | 100 experiments were completed offering a 100% success rate |
| (Chakraa et al., 2023) | Genetic Algorithm | Centralized | Grid Map | ≤ 4 AMRs and 33 tasks | GA approach offers balance between optimality and execution time compared to MILP |
| (Berndt et al., 2021) | FIFO, EDF and LST | Centralized | Python | 6 AGVs | FIFO performs weakly. EDF and LST performance is improved |
| (J. Qiu et al., 2023) | Ant-Colony | Centralized | Python 3.9 | 20 agents | Method outperforms traditional ACO in terms of convergence speed and solution quality |

Papers based on the application of Exact and Matheuristic Methods provide a variety of algorithms for optimization. This approach is the most common in robot fleet TA and includes two main types of algorithms: MILP and Hungarian algorithm algorithms. Similarly to other methods, Exact and Matheuristic solutions are based on centralized management to find globally optimal solutions. Regarding the most used frameworks in this section, Python is also a suitable language to develop the simulations. Furthermore, most of them are based on large fleets with up to 30 robots using this programming language. Additionally, (Valero et al., 2023) provides a remarkable exception as it includes the development of a real experiment. Finally, the main contribution in this type of methods is based on improving the scalability, due to it is one of the main drawbacks of Exact Methods. All this content can be observed in Table III.

TABLE III
EXACT AND MATHEURISTIC ALGORITHMS OVERVIEW

|  | **Method** | **Centralized / Decentralized** | **Framework** | **Fleet Size** | **Remarkable results** |
|---|---|---|---|---|---|
| (Akkaya & Gökçe, 2022) | MILP | Centralized | IBM OPL CPLEX | 1 AGV and 6 charging stations | No charging required to accomplish the tasks by performing the optimization |
| (Hu et al., 2023) | HDSTA | Centralized | Python 3.6 and FlexSim | ≤ 23 AGVs and 150 tasks | Reduction of 81.92% in computation time compared to HALNS |
| (Boccia et al., 2023) | 3S-MHA (MILP based) | Centralized | Python | From 2 to 10 AGVs and ≤ 200 tasks | High scalability with a gap of just 1% while increasing the fleet size |
| (Bergmann et al., 2021) | MIQP | Centralized | C++ (D. Gyulai et al., 2020) | ≤ 17 AGVs and 200 machines | Improved maxium utilization of the AGVs |
| (Chatzisavvas et al., 2022) | Hungarian | Centralized | Pygame (JSON and Python) | ≤ 6 AMRs and 6 tasks | Simulataneous TA and PP |
| (A. Samiei & L. Sun, 2024) | Hungarian | Decentralized | Montecarlo Simulation (language not defined) | ≤ 30 agents and 300 tasks | DMHCBA outperforms the other algorithms in terms of cost and converging time |
| (Msala et al., 2023) | Hungarian | Centralized | MATLAB | ≤ 30 AMRs and 300 tasks | Reduction in the cost to calculate the optimal solution |
| (Singh et al., 2022) | ALNS- based and MILP | Centralized | Python v3.6 and Gurobi package | ≤ 9 AGVs and 60 tasks | MILP better for small-scale scenarios. ALNS outperforms in terms of scalability |
| (Valero et al., 2023) | Fuzzy Optimization | Decentralized | Python | Simulations: 2 AMRs and 4 tasks. Real experiments: 5 AMRs and 6 tasks | Obtention of better results focusing on defining task urgency |

With reference to the different works based on the usage of MBA algorithms, these papers provide a different perspective in terms of architecture. Most of them apply decentralized management due to the application of auction methods, where each robot must calculate its own bids. Considering the different frameworks used, MBA simulations are developed mainly with Python, but also using Robot Operating System (ROS). This aspect provides a clear advantage due to its capabilities to be exported to real robots in the future. Besides, after analyzing the different quantities of robots used in the simulations, most of them are based on robot fleets with up to 6 robots. However, (Teck et al., 2023) provides a more analytical study with a larger fleet (up to 48 robots). Referring to the main contributions that can be extracted, MBA algorithms have provided positive results mainly in terms of execution time and scalability, in exchange of not achieving the most optimal solutions. The results are displayed in Table IV.

TABLE IV
MARKET-BASED APPROACH ALGORITHMS OVERVIEW

| | Method | Centralized / Decentralized | Framework | Fleet Size | Remarkable results |
|---|---|---|---|---|---|
| (De Ryck et al., 2021) | Auction | Decentralized | Pseudocode (not defined) | ≤ 4 AGVs | Less charging time but with higher computational cost |
| (Tavares et al., 2023) | Auction | Decentralized | ROS Noetic, Python3, Gazebo. | ≤ 6 AMRs and ≤200 tasks | More optimal in terms of task execution time compared to other solutions |
| (Teck et al., 2023) | Auction | Decentralized | Python 3.8 | From 50 to 1000 orders, from 3 to 48 AMRs | Regret method has better scalability in terms of number of tasks. LKH-3 gets better optimal solutions despite consuming more time |
| (Galati et al., 2023) | Auction Supervised | Centralized | Pytho ROS node and C++ with Open Motion Planning Library (OMPL) | 3 AMRs, 3 different tasks | Solution provided is more efficient than MILP and heuristic approaches |

Table V provides an overview of publications that use AI methods in TA applications. These types of algorithms compound a current trending method, mainly in terms of RL. The different works analyzed are focused on centralized management. This architecture is influenced by the computational cost of RL applications. In terms of frameworks used for the development, there is not a remarkable environment due to the novelty of the approach. However, Python, ROS and Matlab provide the main picture in this aspect. Regarding the number of robots used during the simulations, the most ambitious simulations make usage of 30 robots (like other types of algorithms). Furthermore, the results show that RL provides reduction of distance traveled in exchange of high computational cost.

TABLE V
ARTIFICIAL INTELLIGENCE BASED ALGORITHMS OVERVIEW

| | Method | Centralized / Decentralized | Framework | Fleet Size | Remarkable results |
|---|---|---|---|---|---|
| (Lei et al., 2023) | Neural Network | Centralized | Based on datasets (language not defined) | 22 AMRs devided into 4 teams | Reduction of distance traveled comparing to Hungarian optimization method |
| (Shibata et al., 2023) | Reinforcement Learning | Hybrid | Python and OpenAI gym | ≤ 6 AMRs | Reduces the transport time while moving objects with different weights |
| (Perumaal Subramanian & Kumar Chandrasekar, 2024) | Reinforcement Learning | Centralized | Matlab 2022a | ≤ 30 AMRs and 12 picking stations | 26% reduction in the maximum distance traveled by the robots. The algorithm outperforms soft computing techniques by 44% in terms of computation time |
| (Malus et al., 2020) | Reinforcement Learning | Centralized | RLib, ROS, Gazebo | 5 AMRs | Comparison provided to show improvements in terms of learning while performing more iterations |

# IV. CONCLUSION

TA problem is presented in this paper, providing the main algorithms that are currently used to solve this optimization problem. Furthermore, this work shows many examples of recent publications that present applications of these algorithms to solve TA. The analysis includes the review of innovative AI based methods which provides a new trend on this field. Additionally, different tables are added to display differences and similarities between the developed simulations.

Moreover, some conclusions can be extracted from the analysis. Firstly, it is necessary to remark the lack of applications in dynamic environments. The usage of simplification during the simulations boosts their development, although makes more difficult later applications in real environments. Another important related aspect is the absence of real experiment during the last years. Therefore, both conclusions represent the necessity to advance towards the development of more real simulations that consider human-sharing environments and other dynamic obstacles, in order to provide experiments in real scenarios.